%% file: sample-sigconf.tex
\documentclass[sigconf]{acmart}
\AtBeginDocument{%
  }

\setcopyright{acmlicensed}
\copyrightyear{2018}
\acmYear{2018}
\acmDOI{XXXXXXX.XXXXXXX}
\acmConference[Conference acronym 'XX]{Make sure to enter the correct
  conference title from your rights confirmation email}{June 03--05,
  2018}{Woodstock, NY}
\acmISBN{978-1-4503-XXXX-X/2018/06}

\newcommand{\name}{EmbedFilter}

\input{math_commands}

\usepackage{algorithm}       
\usepackage{algpseudocode}   
\usepackage{amsmath}         
\usepackage{bm}              

\usepackage{booktabs}    
\usepackage{tabularx}    
\usepackage[table,svgnames]{xcolor} 
\usepackage{graphicx}    
\usepackage{amsmath}     

\usepackage{listings}
\usepackage{caption} 

\definecolor{gray}{rgb}{0.5,0.5,0.5}

\usepackage[most]{tcolorbox}

\usepackage{tikz}

\newcommand{\circledNum}[1]{%
  \tikz[baseline, anchor=base]{\node[draw,circle,inner sep=0.1pt, font=\fontsize{7.5}{10}\selectfont] {#1};}%
}

\begin{document}

\title{Your UnEmbedding Matrix is Secretly a Feature Lens for Text Embeddings}

\author{Songhao Wu}
\orcid{0009-0007-3697-7082}
\authornote{These authors contributed equally. 
Songhao Wu discovered the core phenomenon, provided the core implementation  and led the writing. Zhongxin Chen refined the code, conducted the experiments and provided Songhao Wu with valuable insights.
}
\affiliation{%
  \institution{Gaoling School of Artificial Intelligence, Renmin University of China}
  \city{Beijing}
  \country{China}
}
\email{songhaowu@ruc.edu.cn}

\author{Zhongxin Chen}
\orcid{0009-0004-0735-2196}
\authornotemark[1]
\affiliation{%
  \institution{Gaoling School of Artificial Intelligence, Renmin University of China}
  \city{Beijing}
  \country{China}
}
\email{chenzhongxin@ruc.edu.cn}

\author{Yuxuan Liu}
\orcid{0009-0007-2154-7215}
\affiliation{%
  \institution{Gaoling School of Artificial Intelligence, Renmin University of China}
  \city{Beijing}
  \country{China}
}
\email{yuxuanliu@ruc.edu.cn}

\author{Heng Cui}
\orcid{}
\affiliation{%
  \institution{Lenovo Group Limited}
  \city{Beijing}
  \country{China}
}
\email{cuiheng3@lenovo.com}

\author{Cong Li}
\orcid{}
\affiliation{%
  \institution{Lenovo Group Limited}
  \city{Beijing}
  \country{China}
}
\email{licong17@lenovo.com}

\author{Rui Yan}
\orcid{0000-0002-3356-6823}
\affiliation{%
  \institution{Wuhan University}
  \city{Wuhan}
  \country{China}
}
\authornote{Corresponding Author.}
\email{rui.yan@whu.edu.cn}

\renewcommand{\shortauthors}{Songhao Wu et al.}

\begin{abstract}
Large language models exhibit impressive zero-shot capabilities across a wide range of downstream tasks.
However, they struggle to function as off-the-shelf embedding models, leading to suboptimal performance on massive text embedding benchmarks.
In this paper, we identify a potential cause underlying this deficiency.
Our motivation stems from an unexpected observation: text embeddings tend to align with frequent but uninformative tokens when projected onto the vocabulary space. 
We argue that this excessive expression of high-frequency tokens suppresses the model's ability to capture nuanced semantics.
To address this, we introduce \name, a simple linear transformation designed to refine text embeddings derived from LLMs directly.
Specifically, we uncover that the unembedding matrix within LLMs encodes a latent space that is actively writing these frequent tokens into embedding space.
By filtering out this subspace, \name\ suppress the influence of high-frequency tokens, thereby enhancing semantic representations.
As a compelling byproduct, this enables an inherent dimensionality reduction, lowering index storage and speedup retrieval while fully preserving the refined embedding quality.
Our experiments across multiple LLM backbones demonstrate that LLMs equipped with \name\ achieve superior  zero-shot downstream performance even with significantly reduced embedding dimensions.
We hope our findings provide deeper insights into the mechanisms of LLM-based representations and inspire more principled designs to improve text embeddings training.
Our code is available at \url{https://github.com/CentreChen/EmbFilter}.

\end{abstract}



\maketitle

\section{Introduction}
\label{sec:intro}
Large language models~(LLMs) have made significant strides in recent years, demonstrating impressive performance across a wide range of tasks~\cite{r1,llama3,qwen2.5}.
The emergence of zero-shot learning ability helps LLMs address unseen tasks effectively without any additional fine-tuning~\cite{scaling}.
However, recent studies highlight a persistent performance gap of LLMs when deployed as zero-shot text embedding models~\cite{prompteol,moee,llm2vec}.
This deficiency hinders their adoption for text embedding tasks and raises concerns regarding their full efficacy as generalist models in real-world applications.

\begin{figure*}[t]
    \centering
    \includegraphics[width=0.96\linewidth]{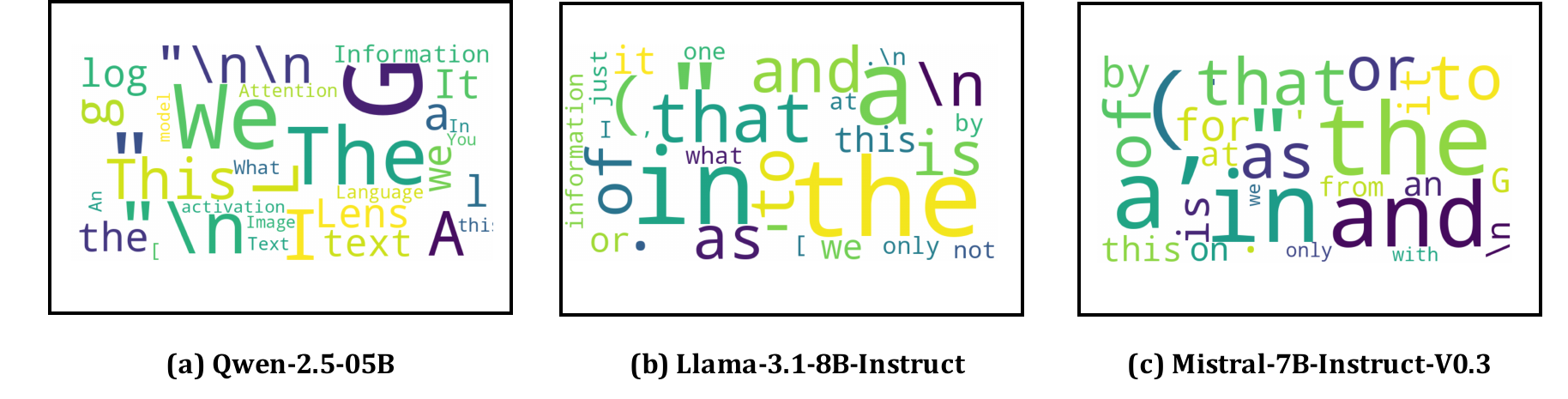}
    \caption{Logit Lens applied to text embeddings from three LLM backbones. Word clouds show the top-aligned tokens with the highest decoding probabilities, which are primarily high-frequency yet semantically uninformative. The input text, encoded by the text embeddings, is given as: \textit{"We call this a `lens' because it is one way of extracting information from GPT's internal activations. I imagine there is other information present in the activations that cannot be understood by looking at logits over tokens. The logit lens show us some of what is going on, not all of it."} This corresponds to the official notation of the logit lens.}
    \label{fig:teaser}
\end{figure*}

To bridge this gap, researchers have explored various attempts to better elicit semantic information from LLMs.
Prompt-engineering methods have been proposed to help extract text embeddings directly from LLMs~\cite{prompteol,echo,metaeol,geneol}.
These approaches are well motivated; however, their improvements are modest and highly sensitive to the choice of the prompt, leading to inconsistent performance across different setups.
Existing approaches are primarily heuristic and fail to resolve the bottleneck that limits LLMs’ ability to capture semantics.
In this paper, we move beyond previous heuristic efforts and seek to provide a mechanistic interpretation for LLMs' suboptimal performance in text embedding tasks. 
Specifically, we identify an unexpected representation collapse: when projected onto the vocabulary space, raw text embeddings from LLMs tend to align with high-frequency tokens that are semantically irrelevant.
Equipped with the Logit Lens tool~\cite{logitlens}, we find that frequent but uninformative tokens disproportionately dominate the highest decoding probabilities of these text embeddings.
This suggests that these hidden representations are biased toward common vocabulary tokens, regardless of the input semantics\footnote{For readers unfamiliar with Logit Lens, please refer to Section~\ref{sec:bg} for further details.}.
As shown in Figure~\ref{fig:teaser}, this phenomenon is observed across different language model families, indicating a universal pattern inherent to LLMs.

We extend our analysis to uncover the underlying drivers of this representation collapse.
Prior studies~\cite{bertflow, ethayarajh-2019-contextual} have established that text embeddings are \textit{anisotropic}: they are confined to a narrow cone rather than being uniformly distributed in the embedding space.
We hypothesize that the centroid of this narrow region corresponds to an ``average” token, which \citet{lv2024fact} describe as the frequency-weighted average embedding over the training corpus. 
This perspective provides a mechanistic rationale for the atypical patterns observed in Logit Lens analyses.
Raw embeddings from LLMs are pulled toward this commonality region, overshadowing their unique semantic features.
By suppressing the contribution of these "average” components, we can mitigate the anisotropy problem and unmask the true semantic representations within LLMs.

We seek to pinpoint the hidden contributor that steer text embeddings towards the "average" token representation. 
To this end, we apply Logit Spectroscopy~\citep{spectral} to a reverse-engineered "average" token, and uncover a latent subspace, which is actively writing these frequent tokens into the embedding space.
We refer to this subspace as the \textit{"edge spectrum"} space, as it is spanned by the right singular vectors with the smallest and largest singular values --- those positioned at the ends of the spectrum.
We find that when the projection of the "average" token onto this subspace is truncated, the logits of these frequent tokens are significantly disrupted.
Section~\ref{sec:interpret} delves into the discovery of the edge spectrum, providing a detailed account of its identification

Leveraging this insight, we show that this subspace can be effectively filtered out via a simple linear transformation, which we term \name.
This transformation is encoded within the parameters of the unembedding matrix and is readily accessible without further training.
Our evaluations across a diverse suite of downstream tasks demonstrate that \name\ acts as a potent post-processing enhancement, delivering steady incremental gains atop existing zero-shot text embedding baselines.
\name\ exhibits strong robustness across various backbone models and experimental configurations while incurring minimal computational overhead.
Beyond performance gains, \name\ naturally lends itself to dimensionality reduction as a distance-preserving transformation.
This reduction lowers indexing overhead and speeds up retrieval, facilitating the practical deployment of LLMs.

To sum up, the contributions of this paper are threefold.

(1) We identify the LLM unembedding matrix as a previously overlooked feature lens to analyze the embedding space.
We reveal that this matrix encodes a latent subspace corresponding to an "average" token and limits the embedding capabilities of LLMs.
We provide an mechanism interpretation that clarifies both the origins and impact of this phenomenon.

(2) We introduce \name, a simple linear transformation that improves the zero-shot text embedding performance of LLMs.
As an efficient post-processing technique, \name\ achieves up to a 14.1\% improvement on MTEB without any training overhead. 
Extensive evaluations across diverse experimental setups further demonstrate its broad applicability.

(3) We demonstrate that \name\ acts as a distance-preserving transformation and enable embedding dimensionality reduction. 
This leads to faster retrieval and lower storage requirements, thereby facilitating the practical deployment of LLMs in large-scale text embedding applications.

\section{Background}
\label{sec:bg}
To establish the background for \name\, we first review the fundamentals of embedding extraction and introduce the mechanistic interpretability tools used throughout our analysis.

\subsection{Text Embedding Paradigm}
We first formulate the standard process of LLM-based text embedding extraction. 
Our objective is to transform sentence \(\mX\) into a dense vector \( \vh \in \mathbb{R}^d\), such that the similarity between these vectors can reflect their semantic similarity.
Given an input sentence \(\mX=\left[x_1,x_2,\dots,x_L\right]\), its embedding \( \vh \) is obtained by passing \(\mX\) through an LLM backbone, followed by a pooling strategy \(\operatorname{P}\):
{\Large \[
\vh \; = \; \operatorname{P} \left (\,\operatorname{LLM}\,(\left[\,x_1,x_2,\dots,x_L\,\right]) \,\right),
\]}
where \( \operatorname{P} \) aggregates the final layer outputs from \(\textrm{LLM}\) into a \(d\)-dimensional representation \( \vh \).
Typically, the unembedding matrix is conceptually designed to map these hidden states back to the vocabulary space for token prediction.
We contend that this module has been overlooked in the context of traditional text embedding extraction and can be exploited to enhance embeddings qualities.

\subsection{Text Embeddings with Prompt Engineering}

Many studies have explored improving the performance of LLMs on text embedding tasks through prompt engineering. 
Here, we provide a brief overview of two well-established baselines:

\textbf{PromptEOL}~\cite{prompteol} finds that a "one word limitation" template can help better condense semantics into the hidden state, thereby enhancing the representation of LLM-derived embeddings.

\textbf{ECHO}~\cite{echo} suggests that causal attention in LLMs is a bottleneck, as earlier tokens cannot access future context. 
To mitigate this, they duplicate the input and extract embeddings from the second occurrence, incurring overhead from the increased input size.

More sophisticated prompt-engineering methods have been proposed~\cite{metaeol,geneol}; however, these often necessitate intricate pipeline designs and incur substantial computational overhead. While our primary experiments focus on the aforementioned baselines, we provide a broader discussion and evaluation of these more complex strategies in our supplementary analysis.

\subsection{Mechanistic Interpretability Tools}
We provide an overview of two interpretability tools --- Logit Lens~\cite{logitlens} and Logit Spectroscopy~\cite{spectral} --- which facilitate the identification of edge spectrum subspace and inspire the design of \name.

\textbf{Logit Lens}~\cite{logitlens} represents a cornerstone of mechanistic interpretability research. 
Its central premise is to project a model’s intermediate representations directly into the vocabulary space.
By analyzing the resulting changes in these logits, researchers can discern how specific intermediate activations shape the final predictions, thereby gaining insights into the model's internal processing logic.
Building on this framework, \citet{nie2024text}~apply the Logit Lens tool to text embeddings and find that these embeddings can align with certain keywords from the input texts.

To further dissect the semantic properties of different embedding subspaces, 
\textbf{Logit Spectroscopy}~\citep{spectral} extends Logit Lens by projecting intermediate representations onto spectral components of model’s weight matrices.
Let \(\mW_\mathcal{U}\) be the unembedding matrix of the LLM. 
Its singular value decomposition can be formulated as:
{ \Large \begin{equation*}
    \label{eq:svd}
    \mW_\mathcal{U} \; = \; \mU \, \Sigma \, \mV^\top,
\end{equation*}}
where \(\bm{W_\mathcal{U}} \in \mathbb{R}^{\left | \mathcal{V}\right | \times d}\), with \(d\) representing the hidden-state dimension and \(|\mathcal{V}|\) the vocabulary size.
For an arbitrary dimension $i \in \{0, \dots, d-1\}$, Logit Spectroscopy introduces a filter \(\bm{\Psi_i}\) that removes the projection of \(\vh\) onto the \(i\)-th right singular vector of \(\mV\). 
Formally, this transformation is defined as:
{\Large \[
    {\bm{\Psi_i}} \; = \; {\mI - \mV_{[i]} \, \mV_{[i]}^{\top}}.
\]}
This operation facilitates the spectral analysis of an LLM's intermediate representations, enabling researchers to measure the contribution of hidden states within different spectral subspaces to the final output.
Section~\ref{sec:interpret} details how we leverage these tools to identify the "edge spectrum" subspace.

\section{Discovery of Edge Spectrum Subspace}
\label{sec:interpret}

\subsection{Motivation}
In this section, we present the preliminaries analyses that motivate the development of \name.
Our investigation is driven by an observed correlation between two key insights:

(1) Raw text embeddings from LLMs are typically anisotropic~\cite{bertflow,whitening}. 
These embeddings are concentrated in a narrow subspace, making them excessively similar to one another; 

(2) LLM-derived embeddings often align with high-frequency tokens that carry little semantics.

These insights lead us to reasonably infer that the narrow subspace is responsible for encoding frequent tokens. 
Consequently, we seek to isolate this subspace and mitigate its impact, thereby alleviating the anisotropy problem in text embedding tasks.
To accomplish this, we first reverse-engineer a "centroid" hidden state representing the “average" token. 
We then perform Logit Spectroscopy on this “average" token, revealing that the edge spectrum subspace drives the emergence of high-frequency tokens. 
We present the technical details of this discovery below.

\subsection{Reverse-Engineering of the Average Token}
We leverage the unembedding matrix, together with word frequencies from training corpus, to reverse-engineer the “average” token.

\subsubsection{Experimental Setup}
We evaluate a diverse set of models, ranging from Qwen-2.5~\cite{qwen2.5} (0.5B) to Mistral-v0.3-Instruct~\cite{mistral7b} (7B) and Llama-3.1 Instruct~\cite{llama3} (8B). By spanning multiple scales and model families, we aim to ensure the universality of our findings.

Since pretraining datasets for these LLMs are not disclosed, we approximate their true word frequency distribution $\vp$ by sampling tokens from open-source corpora. 
Specifically, we select the RedPajama~\cite{redpajama} dataset as our evaluation corpus.
Parallel experiments on alternative corpora produce identical results.
The resulting empirical statistics, denoted as $\hat{\vp}$, serve as a robust proxy for distribution $\vp$ and are adopted throughout the following experiments.

\subsubsection{Reverse-Engineering}
We outline the practical steps for reverse-engineering the "average" token.
For a standard inference step, the unembedding matrix is used to compute the probability distribution over the next token.
Formally, this prediction step is given by:
{\Large \begin{equation*}
    \vq \; = \; \operatorname{Softmax} \left( \, \vh \, \mW_\mathcal{U}^\top \, \right),
\end{equation*}}
where the probability of an arbitrary token \(i\) is given by:
{\Large \begin{equation*}
    \vq_i \; = \; \exp(\vw_i) \; \big/ \; {\textstyle \sum_{j=1}^{|\mathcal{V}|}} \exp(\vw_j).
\end{equation*}}
Given this, the logit $\vw_i$ of the \(i\)-th token is denoted as:
{\Large \[
    \vw_i \; = \; \log(\vq_i) \, + \, \log \sum\nolimits_{j=1}^{|\mathcal{V}|} e^{\vw_j},
\]}
where the second term is a shared bias across all logits, which we redefine as $\vb$.
The logits for decoding \(\vh\) is reformulated as:
{\Large \[
    \vh \, \mW_\mathcal{U}^\top \; = \; \log(\vq) \, + \, \vb.
\]}
By denoting the Moore–Penrose pseudo-inverse~\cite{inverse} of \(\mW_\mathcal{U}^\top\) as \(\mW_\mathcal{U}^{+}\), we can further rewrite the preceding formula as:
{\Large \[
    \vh \; = \; \left( \log(\vq) \, + \, \vb \right) \, \, \mW_\mathcal{U}^{+}.
\]}
We substitute the observed word frequencies \(\hat{\vp}\) and interpret \(\hat{\vh}\) as the "average" token representation over the training corpus.
Formally, the average token embedding is defined as:
{\Large \[
    \hat{\vh} \; = \; \log(\hat{\vp}) \; \mW_{\mathcal{U}}^+ \,,
\]}
where the bias term $\vb$ is omitted for analytical simplicity, since it does not alter the fundamental spectral properties.

\subsubsection{Logit Spectroscopy into Average Token}
Having established the theoretical foundation of Logit Spectroscopy, we now detail its application to the average token.
For each dimension \(i \in \{0, \dots, d-1\}\), we apply a filter \(\bm{\Psi_i}\) to remove the projection of \(\hat{\vh}\) onto the subspace, resulting in the perturbed representation \(\widetilde{\vh}^{(i)}\), defined as:
{\Large \[
    \widetilde{\vh}^{(i)} \; = \; \hat{\vh} \, \left( \mI \, - \, \mV_{[i]} \mV_{[i]}^\top \right).
\]}
We analyze the logit shifts between $\vh$ and $\widetilde{\vh}^{(i)}$ for the $k$ most frequent tokens in the training corpus. 
Let $\mathcal{V}^+$ denote this subset of frequent tokens, formally defined as $\mathcal{V}^+ = \{ j \mid j \in \operatorname{argtopk}(\hat{\vp}) \}$. 
The impact of the filtering operation is then quantified by the cumulative logit differences across these tokens, which is given as:
{\Large \[
    \Delta \pi^{(i)} = \frac{\sum_{j \in \mathcal{V}^+} \left| \widetilde{w}^{(i)}_j - \hat{w}_j \right|}{\sum_{j \in \mathcal{V}^+} \left| \hat{w}_j \right|},
\]}
where \(\hat{\vw_j}\) represents the original logit of the \(j\)-th token, and \(\widetilde{\vw_j}^{(i)}\) denotes the logit after filtering out the subspace spanned by the \(i\)-th right singular vector of \(\mW_{\mathcal{U}}\).  
A higher value of \(\Delta \pi^{\mathrm{(i)}}\) indicates that the \(i\)-th singular subspace exerts a more pronounced influence on the representation of high-frequency tokens.

\begin{figure}[t]
    \centering
    \includegraphics[width=\linewidth]{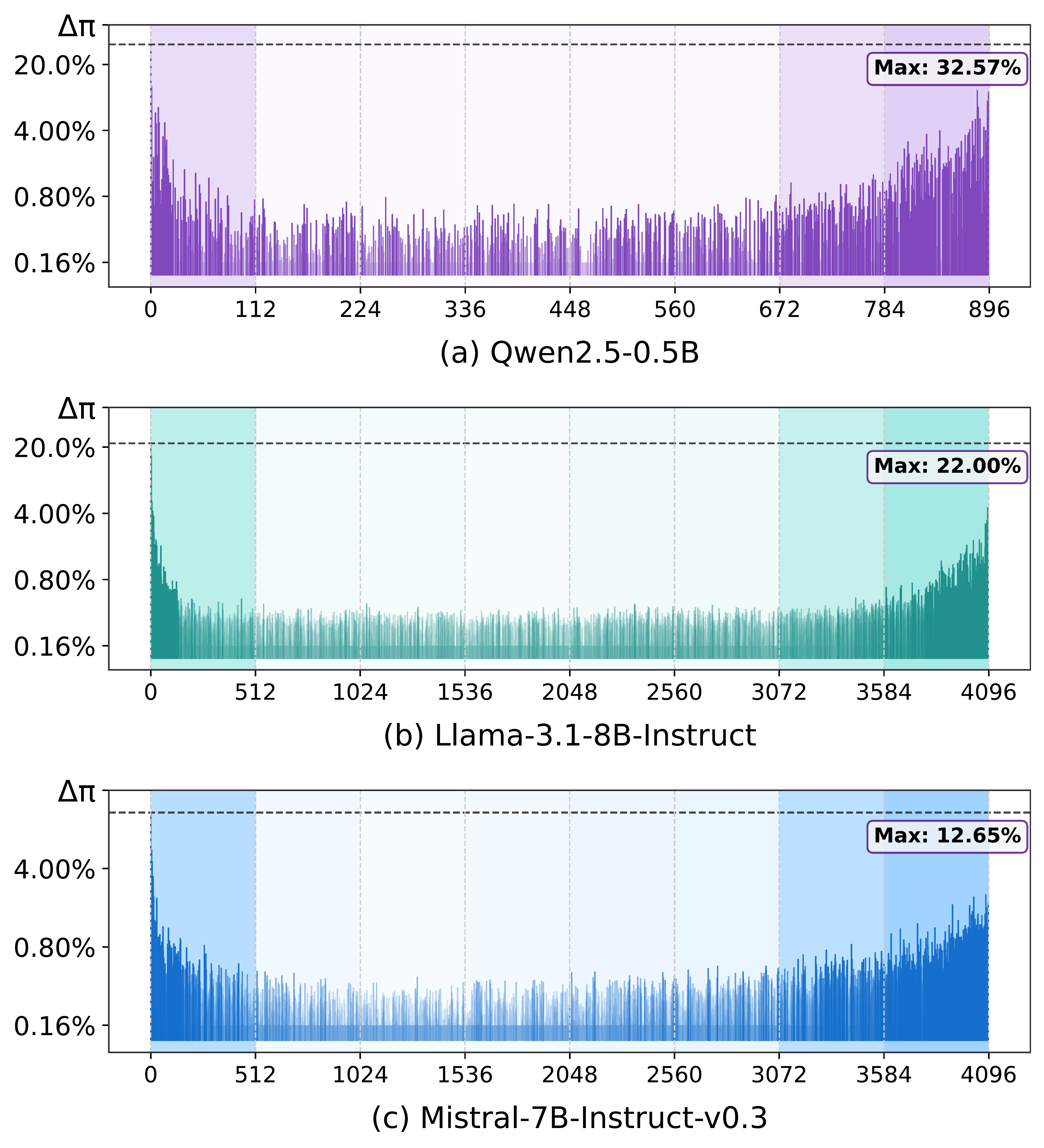}
    \caption{\(\Delta \pi\) distribution for Qwen, Llama and Mistral.}
    \label{fig:edge}
\end{figure}

Figure~\ref{fig:edge} presents the \(\Delta \pi\) values when setting \(k=100\).
As shown, the \(\Delta \pi\) values are significantly larger at the edges of the spectrum, suggesting that the subspaces corresponding to the edge spectrum of LLMs are primarily responsible for encoding high-frequency tokens.
This specific spectral region is precisely what we aimed to identify. 
As demonstrated in the following sections, filtering out this edge spectrum not only suppresses the over-representation of "average" tokens but also enhances the quality of LLM-derived text embeddings.
For comparison, Figure~\ref{fig:infreq} visualizes the influence of different spectral subspaces on the representation of infrequent and randomly sampled tokens.
Notably, the logit differences for infrequent and random tokens exhibit significantly lower sensitivity to the edge spectrum than those for frequent tokens.




\begin{figure*}[t]
    \centering
    \begin{minipage}{0.96\linewidth}
        \centering
        \resizebox{\textwidth}{!}{%
            \begin{footnotesize}
            \begin{tabularx}{0.76\textwidth}{l|*{6}{>{\centering\arraybackslash}X}}
                \toprule
                \multicolumn{1}{c|}{\textbf{Model}} & \multicolumn{6}{c}{ \textbf{Top 6 Tokens From Logit Lens}} \\
                \midrule
                 \textbf{Qwen} & 
                 \textrm{\textbf{G}} &
                 \textrm{\textbf{We}} &
                 \textrm{\textbf{"}} & 
                 \textrm{\textbf{The}} &  
                 \textrm{\textbf{"\textbackslash n}} &
                 \textrm{\textbf{I}} \\
                 \textrm{\quad+~EmbFilter} & 
                 \cellcolor{blue!10} \textrm{\textbf{Language}} &
                 \cellcolor{blue!10} \textrm{\textbf{Lens}} &
                 \cellcolor{blue!10} \textrm{\textbf{anguage}} & 
                  \textrm{\textbf{eca}} &  
                 \cellcolor{blue!10} \textrm{\textbf{agination}} &
                 \cellcolor{blue!10} \textrm{\textbf{\_Language}} \\
                \midrule
                \textbf{Llama} & 
                \textrm{\textbf{\_the}} &
                \textrm{\textbf{,}} &
                \textrm{\textbf{\_a}} &  
                \textrm{\textbf{\_"}} &
                \textrm{\textbf{\_in}} & 
                \textrm{\textbf{\_that}} \\
                \textrm{\quad+~EmbFilter} & 
                \cellcolor{blue!10} \textrm{\textbf{\_activations}} &
                \cellcolor{blue!10} \textrm{\textbf{\_neur}} &
                 \textrm{\textbf{ambre}} &  
                \cellcolor{blue!10} \textrm{\textbf{\_viewpoints}} &
                \cellcolor{blue!10} \textrm{\textbf{\_representations}} & 
                 \textrm{\textbf{sole}} \\
                \midrule
                \textbf{Mistral} & 
                \textrm{\textbf{,}} &
                \textrm{\textbf{the}} &
                \textrm{\textbf{in}} &
                \textrm{\textbf{a}} &
                \textrm{\textbf{(}} &
                \textrm{\textbf{and}} \\
                \textrm{\quad+~EmbFilter} & 
                \cellcolor{blue!10} \textrm{\textbf{hidden}} &
                \cellcolor{blue!10} \textrm{\textbf{activation}} &
                \cellcolor{blue!10} \textrm{\textbf{hidden}} &
                \cellcolor{blue!10} \textrm{\textbf{Hidden}} &
                \cellcolor{blue!10} \textrm{\textbf{lens}} &
                \cellcolor{blue!10} \textrm{\textbf{activ}} \\
                \bottomrule
            \end{tabularx}%
        \end{footnotesize}
        }
        \label{tab:teaser}
    \end{minipage}
    \vspace{0.2cm} 
    \begin{minipage}{0.96\linewidth}
        \centering
        \includegraphics[width=\linewidth]{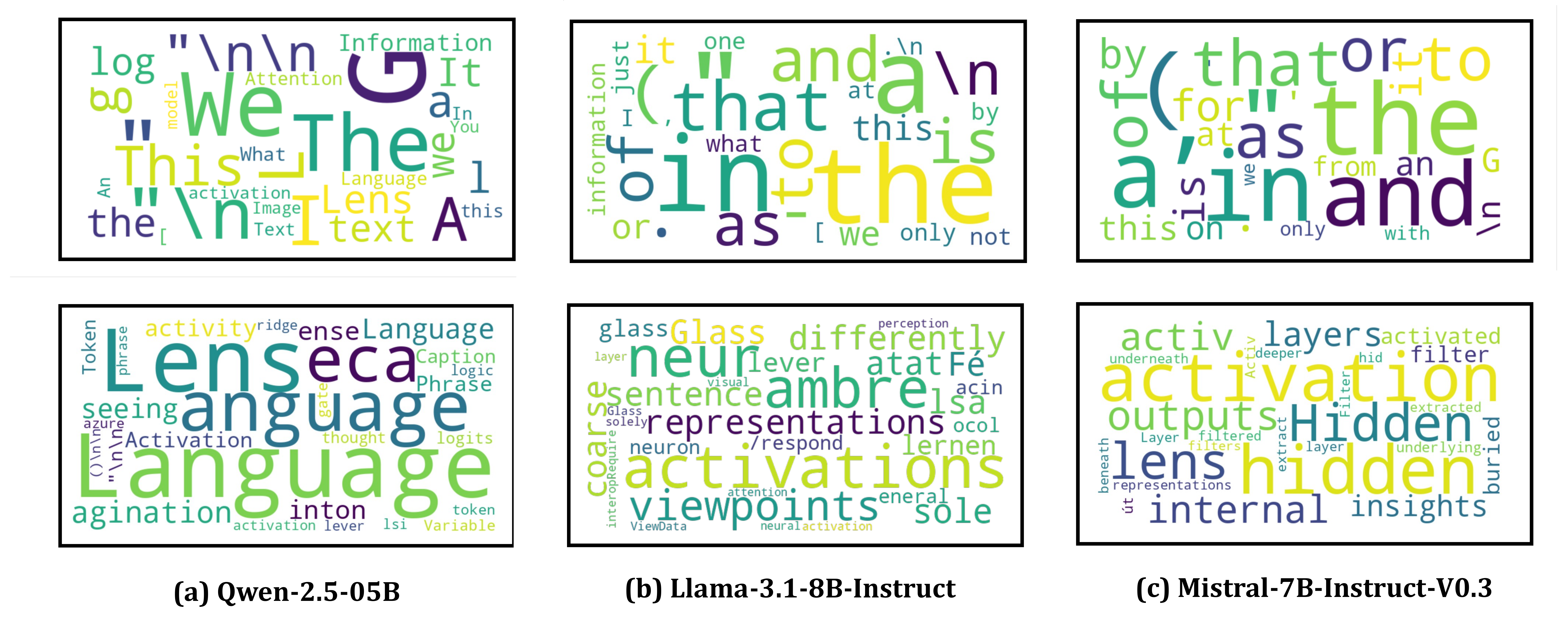}
    \end{minipage}
    \vskip -0.05in
    \caption{
    Re-running logit lens analysis in Section~\ref{sec:intro} with text embeddings refined by \name. 
    Top-6 tokens from logit lens are displayed, with colored entries indicate tokens that have literal connections with the input text. 
    \name\ suppresses the expression of frequent tokens and enhances the semantic richness of text embeddings.
    }
    \label{fig:cmp}
\end{figure*}

\section{Text embedding with \name}
Building on our preliminary insights, we propose \name, a simple linear transformation to filter out the edge spectrum subspace.
This section provides an overview of the \name~workflow.
Additionally, we present a dimensionality reduction approach based on \name\ to highlight its efficiency.

\begin{table*}[t]
    \centering
    \small
    \caption{
    Performance of \name\ across MTEB tasks. 
    \(\tau\) controls dimensionality reduction, scaling the output dimensionality to \(1/\tau\) of the original size.
    Colored entries highlight improvements over the vanilla baseline, while bold text mark the best results within each setup.
    Parenthetical values indicate the performance gain of \name~compared to its baseline.
    }
    \vskip -0.05in
    \label{tab:main_results}
    \resizebox{0.88\textwidth}{!}{
        \begin{tabular}{lcccccccc}
            \toprule
             & \textbf{\textrm{STS.}} & \textbf{\textrm{Class.}} & \textbf{\textrm{Cluster.}} & \textbf{\textrm{PairClass.}} &  \textbf{\textrm{Rerank.}} &  \textbf{\textrm{Retr.}} & \textbf{\textrm{Sum.}} & \textbf{\textrm{Avg.~$\uparrow$}} \\
             \cmidrule(l{2pt}r{2pt}){2-8} \cmidrule(l{1pt}r{1pt}){9-9}
            \textrm{\textbf{Num. Datasets~(\(\rightarrow\))}} & \textbf{10} & \textbf{12} & \textbf{11} & \textbf{3} & \textbf{4} & \textbf{8} & \textbf{1} & \textbf{49} \\
            \midrule
            \multicolumn{1}{c}{} & \multicolumn{7}{c}{\textbf{\textit{\textrm{Qwen2.5-0.5B}}}} \\
            \midrule 
            \textbf{\textrm{PromptEOL}} & 63.04 & 69.20 & 34.91 & 55.15 & 49.33 & 27.31 & 27.30 & 50.07 \\ 
            \cmidrule(l{1pt}r{1pt}){1-9}
            \quad \textbf{\textrm{+~EmbFilter~(\(\mathbf{\tau}=2\))}} & \cellcolor{MediumPurple!15}  \textbf{69.48} & \cellcolor{MediumPurple!15} \textbf{70.32} & \cellcolor{MediumPurple!15} \textbf{39.20} & \cellcolor{MediumPurple!15} \textbf{64.72} & \cellcolor{MediumPurple!15} \textbf{51.28} & \cellcolor{MediumPurple!15} \textbf{34.73} & 27.12 & \cellcolor{MediumPurple!15}  \textbf{54.57~(+9.0\%)}  \\
            \quad \textbf{\textrm{+~EmbFilter~(\(\mathbf{\tau}=4\))}} & \cellcolor{MediumPurple!15} 68.57 & 68.92 & \cellcolor{MediumPurple!15} 38.24 & \cellcolor{MediumPurple!15} 64.54 & \cellcolor{MediumPurple!15} 50.62 & \cellcolor{MediumPurple!15} 32.85 & \cellcolor{MediumPurple!15} 27.67 & \cellcolor{MediumPurple!15} 53.47~(+6.8\%)  \\
            \quad \textbf{\textrm{+~EmbFilter~(\(\mathbf{\tau}=8\))}} & \cellcolor{MediumPurple!15} 68.03 & 66.07 & \cellcolor{MediumPurple!15} 35.50 & \cellcolor{MediumPurple!15} 63.57 & \cellcolor{MediumPurple!15} 49.70 & \cellcolor{MediumPurple!15} 29.82 & \cellcolor{MediumPurple!15} \textbf{28.37} & \cellcolor{MediumPurple!15} 51.43~(+2.7\%)  \\
            \cmidrule(l{4pt}r{4pt}){1-9}
            \textbf{\textrm{ECHO}} & 63.98 & 64.86 & 30.16 & 55.54 & 42.80 & 18.15 & 22.78 & 46.03 \\
            \cmidrule(l{1pt}r{1pt}){1-9}
            \quad \textbf{\textrm{+~EmbFilter~(\(\mathbf{\tau}=2\))}} & \cellcolor{MediumPurple!15} \textbf{70.77} & \cellcolor{MediumPurple!15} \textbf{67.37} & \cellcolor{MediumPurple!15} \textbf{36.94} & \cellcolor{MediumPurple!15} \textbf{66.35} & \cellcolor{MediumPurple!15} \textbf{46.59} & \cellcolor{MediumPurple!15} \textbf{29.65} & \cellcolor{MediumPurple!15} 29.73 & \cellcolor{MediumPurple!15} \textbf{52.55~(+14.1\%)} \\
            \quad \textbf{\textrm{+~EmbFilter~(\(\mathbf{\tau}=4\))}} & \cellcolor{MediumPurple!15} 69.64 & \cellcolor{MediumPurple!15} 65.59 & \cellcolor{MediumPurple!15} 36.17 & \cellcolor{MediumPurple!15} 65.33 & \cellcolor{MediumPurple!15} 46.40 & \cellcolor{MediumPurple!15} 28.61 & \cellcolor{MediumPurple!15} \textbf{31.65} & \cellcolor{MediumPurple!15} 51.50~(+11.9\%) \\
            \quad \textbf{\textrm{+~EmbFilter~(\(\mathbf{\tau}=8\))}} & \cellcolor{MediumPurple!15} 68.81 & 61.91 & \cellcolor{MediumPurple!15} 34.80 & \cellcolor{MediumPurple!15} 63.57 & \cellcolor{MediumPurple!15} 46.13 & \cellcolor{MediumPurple!15} 25.42 & \cellcolor{MediumPurple!15} 29.79 & \cellcolor{MediumPurple!15} 49.43~(+7.4\%) \\
            \midrule
            \multicolumn{1}{c}{} & \multicolumn{7}{c}{\textbf{\textit{\textrm{Llama-3.1-8B-Instruct}}}} \\
            \midrule
            \textbf{\textrm{PromptEOL}} & 75.19 & 73.39 & 39.30 & 64.22 & 53.67 & 25.45 & 25.49 & 55.13   \\
            \cmidrule(l{1pt}r{1pt}){1-9}
            \quad \textbf{\textrm{+~EmbFilter~(\(\mathbf{\tau}=2\))}} & \cellcolor{MediumTurquoise!15} \textbf{76.66} & \cellcolor{MediumTurquoise!15} \textbf{73.78} & \cellcolor{MediumTurquoise!15} \textbf{40.67} & \cellcolor{MediumTurquoise!15} \textbf{66.64} & \cellcolor{MediumTurquoise!15} \textbf{54.68} & \cellcolor{MediumTurquoise!15} 29.69 & \cellcolor{MediumTurquoise!15} 27.39 & \cellcolor{MediumTurquoise!15} \textbf{56.79~(+3.0\%) }\\
            \quad \textbf{\textrm{+~EmbFilter~(\(\mathbf{\tau}=4\))}} & \cellcolor{MediumTurquoise!15} 76.63 & \cellcolor{MediumTurquoise!15} 73.73 & \cellcolor{MediumTurquoise!15} 40.57 & \cellcolor{MediumTurquoise!15} 66.63 & \cellcolor{MediumTurquoise!15} 54.65 & \cellcolor{MediumTurquoise!15} \textbf{29.86} & \cellcolor{MediumTurquoise!15} 27.51 & \cellcolor{MediumTurquoise!15} 56.78~(+3.0\%) \\
            \quad \textbf{\textrm{+~EmbFilter~(\(\mathbf{\tau}=8\))}} & \cellcolor{MediumTurquoise!15} 76.33 & 73.10 & \cellcolor{MediumTurquoise!15} 40.32 & \cellcolor{MediumTurquoise!15} 66.41 & \cellcolor{MediumTurquoise!15} 54.41 & \cellcolor{MediumTurquoise!15} 29.70 & \cellcolor{MediumTurquoise!15} \textbf{27.93} & \cellcolor{MediumTurquoise!15} 56.46~(+2.4\%) \\
            \cmidrule(l{4pt}r{4pt}){1-9}
            \textbf{\textrm{ECHO}} & 70.43 & 68.80 & 38.89 & 66.98 & 49.26 & 30.14 & 25.41 & 53.52 \\
            \cmidrule(l{1pt}r{1pt}){1-9}
            \quad \textbf{\textrm{+~EmbFilter~(\(\mathbf{\tau}=2\))}} & \cellcolor{MediumTurquoise!15} \textbf{74.41} & \cellcolor{MediumTurquoise!15} \textbf{69.77} & \cellcolor{MediumTurquoise!15} \textbf{42.64} & \cellcolor{MediumTurquoise!15} \textbf{73.98} & \cellcolor{MediumTurquoise!15} \textbf{53.15} & \cellcolor{MediumTurquoise!15} \textbf{39.21} & \cellcolor{MediumTurquoise!15} 28.46 & \cellcolor{MediumTurquoise!15} \textbf{57.70~(+7.8\%)} \\
            \quad \textbf{\textrm{+~EmbFilter~(\(\mathbf{\tau}=4\))}}  & \cellcolor{MediumTurquoise!15} 74.20 & \cellcolor{MediumTurquoise!15} 69.13 & \cellcolor{MediumTurquoise!15} 42.28 & \cellcolor{MediumTurquoise!15} 73.94 & \cellcolor{MediumTurquoise!15} 53.07 & \cellcolor{MediumTurquoise!15} 38.64 & \cellcolor{MediumTurquoise!15} \textbf{28.97} & \cellcolor{MediumTurquoise!15} 57.32~(+7.1\%) \\
            \quad \textbf{\textrm{+~EmbFilter~(\(\mathbf{\tau}=8\))}} & \cellcolor{MediumTurquoise!15} 74.05 & 67.50 & \cellcolor{MediumTurquoise!15} 41.88 & \cellcolor{MediumTurquoise!15} 73.76 & \cellcolor{MediumTurquoise!15} 52.75 & \cellcolor{MediumTurquoise!15} 37.75 & \cellcolor{MediumTurquoise!15} 28.58 & \cellcolor{MediumTurquoise!15} 56.61~(+5.8\%) \\
            \midrule
            \multicolumn{1}{c}{} & \multicolumn{7}{c}{\textbf{\textit{\textrm{Mistral-7B-Instruct-v0.3}}}} \\
            \midrule
            \textbf{\textrm{PromptEOL}} & 64.15 & \textbf{71.26} & 33.40 & 58.51 & 48.10 & 20.91 & 24.72 & 49.47   \\
            \cmidrule(l{1pt}r{1pt}){1-9}
            \quad \textbf{\textrm{+~EmbFilter~(\(\mathbf{\tau}=2\))}} & \cellcolor{DodgerBlue!15} 66.59 & 71.17 & \cellcolor{DodgerBlue!15} 36.16 & \cellcolor{DodgerBlue!15} 62.07 & \cellcolor{DodgerBlue!15} 49.63 & \cellcolor{DodgerBlue!15} 24.59 & 24.33 & \cellcolor{DodgerBlue!15} 51.50~(+4.1\%) \\
            \quad \textbf{\textrm{+~EmbFilter~(\(\mathbf{\tau}=4\))}} & \cellcolor{DodgerBlue!15} 67.55 & 70.92 & \cellcolor{DodgerBlue!15} 37.41 & \cellcolor{DodgerBlue!15} 63.29 & \cellcolor{DodgerBlue!15} 50.11 & \cellcolor{DodgerBlue!15} \textbf{25.97} & 24.66 & \cellcolor{DodgerBlue!15} 52.26~(+5.6\%) \\
            \quad \textbf{\textrm{+~EmbFilter~(\(\mathbf{\tau}=8\))}} & \cellcolor{DodgerBlue!15} \textbf{68.11} & 70.07 & \cellcolor{DodgerBlue!15} \textbf{38.04} & \cellcolor{DodgerBlue!15} \textbf{63.67} & \cellcolor{DodgerBlue!15} \textbf{50.20} & \cellcolor{DodgerBlue!15} 25.92 & \cellcolor{DodgerBlue!15} \textbf{25.79} &  \cellcolor{DodgerBlue!15} \textbf{52.35~(+5.8\%) }\\
            \cmidrule(l{4pt}r{4pt}){1-9}
            \textbf{\textrm{ECHO}}  & 72.81 & 71.60 & 32.42 & 71.48 & 47.56 & 28.37 & 31.49 & 53.21 \\
            \cmidrule(l{1pt}r{1pt}){1-9}
            \quad \textbf{\textrm{+~EmbFilter~(\(\mathbf{\tau}=2\))}} & \cellcolor{DodgerBlue!15} 74.66 & \cellcolor{DodgerBlue!15} \textbf{71.79} & \cellcolor{DodgerBlue!15} 36.14 & \cellcolor{DodgerBlue!15} \textbf{74.96} & \cellcolor{DodgerBlue!15} 51.66 & \cellcolor{DodgerBlue!15} 35.03 & 31.23 & \cellcolor{DodgerBlue!15} 56.10~(+5.4\%) \\
            \quad \textbf{\textrm{+~EmbFilter~(\(\mathbf{\tau}=4\))}}  & \cellcolor{DodgerBlue!15} 74.85 & 71.05 & \cellcolor{DodgerBlue!15} \textbf{37.07} & \cellcolor{DodgerBlue!15} 74.91 & \cellcolor{DodgerBlue!15} \textbf{51.87} & \cellcolor{DodgerBlue!15} \textbf{35.49} & 31.14 & \cellcolor{DodgerBlue!15} \textbf{56.25~(+5.7\%) }\\
            \quad \textbf{\textrm{+~EmbFilter~(\(\mathbf{\tau}=8\))}} & \cellcolor{DodgerBlue!15} \textbf{74.86} & 70.00 & \cellcolor{DodgerBlue!15} 36.92 & \cellcolor{DodgerBlue!15} 74.29 & \cellcolor{DodgerBlue!15} 51.71 & \cellcolor{DodgerBlue!15} 34.91 & \cellcolor{DodgerBlue!15} \textbf{31.56} & \cellcolor{DodgerBlue!15} 55.82~(+4.9\%) \\

            \bottomrule
        \end{tabular}
    }
\end{table*}

\subsection{Methodology Formulation of \name.} 
We introduce the Bulk Spectrum Transformation ($\bm{\Phi}_r$), to filter out the edge spectrum space of raw LLM-derived text embeddings.
By excluding the right singular vectors associated with both the largest and smallest singular values, we construct \(\bm\Phi_r\) from the remaining mid-range singular components. 
We hypothesize that this "bulk" of the spectrum suppresses the influence of non-semantic tokens, thereby enabling a more effective capture of core semantics within the embedding space.
Formally, the matrix \(\bm\Phi_r\) is defined as:
{\Large \begin{equation*}
    \label{eq:filter}
    \bm\Phi_{\tau} \; = \; \mV{\left[l_\tau:r_\tau\right]
    \, \mV{\left[l_\tau:r_\tau\right]}^{\top}},
\end{equation*}}
where \(\tau\) is a predefined filtering ratio, with \(l_\tau\) and \(r_\tau\) denoting the start and end indices of the columns.
We use this transformation to post-process the existing embeddings \(\left\{\ve_i\right\}_{i=1}^{N}\),
and map them into refined representations \(\widetilde{\ve_i}\) optimized for downstream tasks:
{\Large \[
    \widetilde{\ve_i} \; = \; \ve_i \, \bm{\Phi_\tau}^\top.
\]}
This transformation safely filters out the edge spectrum space while preserving the components in the bulk spectrum. 
Further implementation details can be found in our code repository.
We then use \name\ to refine the text embeddings and re-run the Logit Lens analysis, with the corresponding before-and-after comparisons presented in Figure~\ref{fig:cmp}.

\subsection{Dimensionality Reduction}
Moreover, we observe that text embeddings refined by \name\ facilitate dimensionality reduction for free.
Recall that \(\mV\) represents the right singular vectors of \(\mW_{\mathcal{U}}\).
Since \(\mV\) is an orthogonal matrix, it constitutes, by definition, a distance-preserving transformation.
Given that, for any \(\vx,\vy \in \mathbb{R}^d\), we have the identity:
{\Large \begin{equation}
    \label{eq:identity}
    \|\vx \, \bm{\Phi_\tau}^\top - \vy \, \bm{\Phi_\tau}^\top\|_2 \; = \; \|\vx \,\mV{\left[l_\tau:r_\tau\right]} - \vy \, \mV{\left[l_\tau:r_\tau\right]}\|_2.
\end{equation}}
Given the properties presented in Equation~\ref{eq:identity}, we can replace \(\bm{\Phi_r}^\top\) with \(\mV\left[l_\tau:r_\tau\right]\), which causes no theoretical difference in similarity measurement.
For readers unfamiliar with these properties, we also provide a simple proof of Equation~\ref{eq:identity} in the Appendix~\ref{sec:proof}.

By invoking this identity transformation, we substantially reduce the hidden size of the raw text embeddings. 
This reduction translates to reduced index storage overhead and faster retrieval speeds, as it minimizes both memory bandwidth bottlenecks and distance computation complexity during search. 
Our experimental results in Section~\ref{sec:exp} demonstrate that this approach successfully achieves significant dimensionality reduction while maintaining or even exceeding downstream task performance, thereby achieving improvements in both efficiency and effectiveness simultaneously.

\section{Experiment}
\label{sec:exp}
\subsection{General Setup.}
We evaluate \name's effectiveness on the MTEB benchmark~\cite{mteb}, which includes standard downstream applications for text embeddings such as Semantic Textual Similarity (STS), Classification (Class.), Clustering (Cluster.), and Retrieval (Retr.).
We build our evaluation framework upon the official MTEB implementation and report the standard metrics for each task.
Due to limited computational resources, we evaluate a subset of the retrieval tasks, following the protocols in \cite{llm2vec, moee}.
Detailed descriptions of the experimental configurations and subset selection can be found in Appendix~\ref{sec:main_setup}.
We evaluate \name\ across three backbone LLMs (Qwen, Llama, and Mistral), ensuring comprehensive coverage of mainstream architectures and model scales.

\subsection{Main Results on MTEB.}
Table~\ref{tab:main_results} presents the main experimental results of \name\ on MTEB, configured with both PromptEOL and ECHO.
Specifically, we analyze \name's performance with different filtering ratios to assess its sensitivity.
We have the following observations:

\begin{table*}[t]
    \centering
    \small
    \caption{Performance of \name\ on MTEB via MetaEOL prompting.}
    \vskip -0.05in
    \label{tab:metaeol}
    \resizebox{0.86\textwidth}{!}{
        \begin{tabular}{lcccccccc}
            \toprule
             & \textbf{\textrm{STS.}} & \textbf{\textrm{Class.}} & \textbf{\textrm{Cluster.}} & \textbf{\textrm{PairClass.}} &  \textbf{\textrm{Rerank.}} &  \textbf{\textrm{Retr.}} & \textbf{\textrm{Sum.}} & \textbf{\textrm{Avg.~$\uparrow$}} \\
            \midrule
            \textbf{\textrm{MetaEOL~(Qwen)}} & 67.15 & 71.43 & 33.44 & 69.09 & 50.26 & 28.17 & 29.28 & 52.23 \\ 
            \cmidrule(l{1pt}r{1pt}){1-9}
            \quad \textbf{\textrm{+~EmbFilter~(\(\mathbf{\tau}=2\))}} & \cellcolor{MediumPurple!15}  \textbf{71.27} & \cellcolor{MediumPurple!15}  \textbf{71.69} & \cellcolor{MediumPurple!15} \textbf{37.19} & \cellcolor{MediumPurple!15}  \textbf{72.28} & \cellcolor{MediumPurple!15}  \textbf{51.65} & \cellcolor{MediumPurple!15} \textbf{34.58} & \cellcolor{MediumPurple!15} \textbf{31.83 }& \cellcolor{MediumPurple!15}  \textbf{55.39~(+6.1\%)}  \\
            \quad \textbf{\textrm{+~EmbFilter~(\(\mathbf{\tau}=4\))}} & \cellcolor{MediumPurple!15}  70.54 & 70.33 & \cellcolor{MediumPurple!15}  36.00 & \cellcolor{MediumPurple!15}  71.57 & \cellcolor{MediumPurple!15}  50.82 & \cellcolor{MediumPurple!15}  33.82 & \cellcolor{MediumPurple!15}  30.80 & \cellcolor{MediumPurple!15}  54.38~(+4.1\%)  \\
            \midrule
            \textbf{\textrm{MetaEOL~(Llama)}} & 71.23 & 74.89 & 41.31 & 72.50 & 52.44 & 32.16 & 29.87 & 56.73 \\ 
            \cmidrule(l{1pt}r{1pt}){1-9}
            \quad \textbf{\textrm{+~EmbFilter~(\(\mathbf{\tau}=2\))}}  & \cellcolor{MediumTurquoise!15} \textbf{73.68} & \cellcolor{MediumTurquoise!15} \textbf{75.53} & \cellcolor{MediumTurquoise!15} \textbf{43.15} & \cellcolor{MediumTurquoise!15} \textbf{75.08} & \cellcolor{MediumTurquoise!15} 53.60 & \cellcolor{MediumTurquoise!15} \textbf{36.60} & \cellcolor{MediumTurquoise!15} 30.42 & \cellcolor{MediumTurquoise!15} \textbf{58.79~(+3.6\%)}  \\
            \quad \textbf{\textrm{+~EmbFilter~(\(\mathbf{\tau}=4\))}} & \cellcolor{MediumTurquoise!15} 73.59 & \cellcolor{MediumTurquoise!15} 75.47 & \cellcolor{MediumTurquoise!15} 42.89 & \cellcolor{MediumTurquoise!15} 75.02 & \cellcolor{MediumTurquoise!15} \textbf{53.61} & \cellcolor{MediumTurquoise!15} 36.41 & \cellcolor{MediumTurquoise!15} \textbf{30.62} & \cellcolor{MediumTurquoise!15} 58.67~(+3.6\%)  \\
            \bottomrule
        \end{tabular}
    }
\end{table*}

\begin{table*}[t]
    \centering
    \small
    \caption{Performance of \name\ on STS tasks under the GenEOL framework.}
    \vskip -0.05in
    \label{tab:geneol}
    \resizebox{0.92\textwidth}{!}{
        \begin{tabular}{l|cccccccccc|c}
            \toprule
             & \textbf{STS12} & \textbf{STS13} & \textbf{STS14} & \textbf{STS15} & \textbf{STS16} & \textbf{STS17} & \textbf{STS22} & \textbf{SICK-R} & \textbf{STSB} & \textbf{BIOSSES} & \textbf{Avg. \(\uparrow\)} \\
             \midrule
            \textbf{\textrm{GenEOL}} & 71.36 & 84.89 & 77.29 & 80.94 & 81.17 & 84.21 & 67.72 & 78.19	& 79.23 & 72.27 & 77.73 \\
            \midrule
            \textbf{\textrm{+ \, EmbFilter~(\(\mathbf{\tau}=2\))}} & 71.28	& 85.19	& 77.92 & 	81.60 & 81.87 & 86.08 & 68.51 & 78.89 & 80.14 & 76.38 & \cellcolor{MediumPurple!15}  \textbf{78.39} \\
            \textbf{\textrm{+ \, EmbFilter~(\(\mathbf{\tau}=2\))}} & 70.38	& 84.81	& 77.20 & 	81.00 & 81.22 & 85.15 & 66.99 & 78.31 & 79.31 & 76.42 & \cellcolor{MediumPurple!15}  78.08 \\
            \bottomrule
        \end{tabular}
    }
\end{table*}

(1) \name\ demonstrates notable improvements across all experimental setups, providing strong evidence of its effectiveness and robustness. 
Specifically, \name\ delivers remarkable enhancements over the baselines, achieving up to a 14\% increase in MTEB overall performance.
These performance gains are maintained even when the output embedding size is reduced to only $1/8$ of its original dimension. 
Furthermore, \name\ consistently achieves superior overall performance across all evaluated setups, whereas the prompt-engineering methods exhibits performance fluctuations.
This underscores the generalization capability of \name\ and highlights its potential for integration with a broader spectrum of LLMs.

(2) \name\ introduces only a lightweight linear transformation module, ensuring negligible overhead during the post-processing of large-scale text embeddings.
Additional experimental results in Table~\ref{tab:metaeol} and~\ref{tab:geneol}, demonstrate that \name\ remains highly effective even when integrated into sophisticated prompt-engineering pipelines, such as MetaEOL~\cite{metaeol} and GenEOL~\cite{geneol}.
Unlike these complex frameworks --- which requires iterative calls to powerful commercial LLMs or the aggregation of multiple embeddings for a single sentence ---
\name\ bypasses the heavy computational overhead of these complex extraction framework design, leading to superior downstream performance with higher efficiency.

\subsection{The Effect of Filtering Ratio \(\tau\)}
As aforementioned, we introduce a hyperparameter \(\tau\) to represent the filtering ratio in \name.
Consequently, the dimensionality of text embeddings is reduced to \(1/\tau\) of the original size.
This reduction is critical, as it scales down the index storage to \(1/\tau\) of its previous occupation and theoretically result in \(\tau \times\) speedup in similarity computation.
A larger value of \(\tau\) indicates lower memory usage and faster retrieval speeds, which is especially beneficial in real-world applications.
Based on this, we analyze the impact of \(\tau\) on the performance of \name\.
As shown in Table~\ref{tab:main_results}, \name\ consistently delivers improvement acorss different choices of \(\tau\).
Remarkably, it retains competitive, and in some cases, superior performance on MTEB tasks, even at a high filtering ratio of \(\tau=8\).

Large language models typically have larger hidden sizes, leading to increased storage and computational costs when deployed as embedding models.
By incorporating \name, LLMs can attain improved downstream performance with smaller representation dimensions.
We present the dimensionality reduction performance of Llama-3.1-8B-Instruct with \name~in Table~\ref{tab:tau}.
With the aid of \name, zero-shot LLMs can outperform established, well-trained baselines from the pre-LLM era, such as SimCSE~\cite{simcse} and coCondensor~\cite{cocondensor}, while utilizing smaller representation dimensions.
This advancement enables the direct deployment of LLMs as embedding models in low-resource scenarios.

\begin{table*}[t]
    \centering
    \small
    \caption{Dimensionality reduction performance of Llama with \name~on MTEB. }
    \vskip -0.05in
    \label{tab:tau}
    \resizebox{0.76\textwidth}{!}{
        \begin{tabular}{l|c|ccccccc|c}
            \toprule
            & \textbf{Dim.} & \textbf{\textrm{STS}} & \textbf{\textrm{Class.}} & \textbf{\textrm{Cluster.}} & \textbf{\textrm{PairClass.}} &  \textbf{\textrm{Rerank.}}  &  \textbf{\textrm{Retr.}} & \textbf{\textrm{Sum.}} & \textbf{\textrm{Avg.}~($\uparrow$)} \\
            \midrule
            \textbf{\textrm{SimCSE-BERT-sup}} & \textbf{768} & 79.12 & 67.32 & 33.43 & 74.89 & 47.53 & 26.34 & 31.17 & 53.54 \\
            \textbf{\textrm{coCondenser-msmarco}} & \textbf{768} & 76.47 & 64.71 & 37.64 & 81.74 & 51.85 & 35.14 & 29.50 & 55.48 \\
            \cmidrule(l{2pt}r{2pt}){1-10}
            \textbf{\textrm{PromptEOL}} & \textbf{4096} & 70.43 & 68.80 & 38.89 & 66.98 & 49.26 & 30.14 & 25.41 & 53.52  \\
            \quad \textbf{\textrm{+~EmbFilter~(\(\mathbf{\tau}=8\))}} & \textbf{4096} & 76.33 & 73.10 & 40.32 & 66.41 &  54.41 &  29.70 &  \textbf{27.93} & \cellcolor{MediumTurquoise!15} \textbf{56.46} \\
            \midrule
            \textbf{\textrm{ECHO}} & \textbf{4096} & 70.43 & 68.80 & 38.89 & 66.98 & 49.26 & 30.14 & 25.41 & 53.52 \\
            \textbf{\textrm{\quad+EmbFilter~(\(\mathbf{\tau}=8\))}} & \textbf{512} & 74.05 & 67.50 & 41.88 & 73.76 & 52.75 & 37.75 & 28.58 & \cellcolor{MediumTurquoise!15} \textbf{56.61} \\
            \midrule
            \textbf{\textrm{MetaEOL}} & \textbf{4096} & 71.23 & 74.89 & 41.31 & 72.50 & 52.44 & 32.16 & 29.87 & 56.73 \\ 
            \quad \textbf{\textrm{+~EmbFilter~(\(\mathbf{\tau}=8\))}} & \textbf{512} & 73.49 & 74.74 & 42.47 & 74.55 & 53.39 & 35.89 & 30.72 & \cellcolor{MediumTurquoise!15} \textbf{58.25}  \\  
            \bottomrule
        \end{tabular}
    }
\end{table*}

\begin{table*}[t]
    \small
    \centering
    \caption{Ablation studies of the filtering strategies. Best results are in bold.}
    \vskip -0.05in
    \label{tab:ablation}
    \resizebox{0.68\textwidth}{!}{
        \begin{tabular}{lcccccccc}
            \toprule
             & \textbf{\textrm{STS}} & \textbf{\textrm{Class.}} & \textbf{\textrm{Cluster.}} &\textbf{\textrm{PairClass.}} &  \textbf{\textrm{Rerank.}} &  \textbf{\textrm{Retr.}} & \textbf{\textrm{Sum.}} & \textbf{\textrm{Avg.} } \\
            \midrule
            \textrm{PromptEOL} & 63.04 & 69.20 & 34.91 & 55.15 & 49.33 & 27.31 & 27.30 & 50.07 \\ 
            \quad \textrm{+ EmbFilter} & \textbf{69.48} & 70.32 & \textbf{39.20} & 64.72 & \textbf{51.28} & \textbf{34.73} & 27.12 & \textbf{54.57}  \\
            \midrule
            \circledNum{1} \; \textrm{Truncation} & 62.56 & 68.54 & 33.52 & 54.81 & 48.93 & 25.34 & 27.67 & 49.13 \\
            \circledNum{2} \; \textrm{Random} & 63.27 & 68.29 & 34.15 & 54.55 & 48.81 & 25.03 & 27.03 & 49.27 \\
            \cmidrule(l{2pt}r{2pt}){1-9}
            \circledNum{3} \; \textrm{Dominant} & 60.34 & 66.97 & 33.25 & 51.18 & 48.13 & 22.89 & 27.00 & 47.53 \\
            \circledNum{4} \; \textrm{Secondary} & 67.74 & \textbf{70.49} & 36.28 & 62.71 & 50.27 & 33.17 & \textbf{29.26} & 53.19 \\
            \circledNum{5} \; \textrm{Bulk} & 59.92 & 67.22 & 32.08 & 50.71 & 47.83 & 22.47 & 27.46 & 47.13 \\
            \cmidrule(l{2pt}r{2pt}){1-9}
            \circledNum{6} \; \textrm{Optimal} & 68.52 & 69.97 & 38.63 & \textbf{65.03} & 51.03 & 34.68 & 28.67 & 54.19 \\
            \bottomrule
        \end{tabular}
    }
\end{table*}

\begin{table*}[t]
    \centering
    \caption{MTEB results for \name~and whitening. Best results are highlighted in bold. }
    \vskip -0.05in
    \label{tab:whitening}
    \resizebox{0.78\textwidth}{!}{
        \begin{tabular}{c|c|ccccccc|c}
            \toprule
            & \textbf{Dim.} & \textbf{\textrm{STS}} & \textbf{\textrm{Class.}} & \textbf{\textrm{Cluster.}} & \textbf{\textrm{PairClass.}} &  \textbf{\textrm{Rerank.}}  &  \textbf{\textrm{Retr.}} & \textbf{\textrm{Sum.}} & \textbf{\textrm{Avg.}}~($\uparrow$) \\
            \midrule
            \textbf{\textrm{PromptEOL}} & \textbf{896} & 63.04 & 69.20 & 34.91 & 55.15 & 49.33 & 27.31 & \textbf{27.30} & 50.07 \\
            \midrule
            \textbf{\textrm{EmbFilter}} & \textbf{448} & \textbf{69.48} & \textbf{70.32} & \textbf{39.20} & 64.72 & \textbf{51.28} & \textbf{34.73 }& 27.12 & \textbf{54.57~(+9.0\%)} \\
            \textbf{\textrm{whitening}} & \textbf{448} & 67.18 & 69.62 & 36.03 & \textbf{67.67} & 50.56 & 32.92 & 26.98 & 53.04~(+5.9\%) \\
            \bottomrule
        \end{tabular}
    }
\end{table*}

\subsection{Ablation Studies of Filtering Strategies}
We evaluate various configurations of our filtering strategies to verify the effectiveness of the \name\ design.
Specifically, we conduct a detailed ablation analysis using Qwen2.5-0.5B with PromptEOL and a dimensionality reduction ratio of \(\tau=2\).
The results across these different experimental setups are reported in Table~\ref{tab:ablation}.
We can draw the following conclusions:

(1) The improvement of \name~does not stem from a simple reduction in the dimensionality of text embeddings.
For configuration \circledNum{1}, we truncate the first half of the dimensions from the original text embeddings, following the Matryoshka setup~\cite{kusupati2022matryoshka}.
In configuration \circledNum{2}, we randomly choose half of the dimensions from the original \(d\)-dimensional vector to form the reduced embeddings.
Configuration \circledNum{1} and \circledNum{2} have fewer vector dimensions but still underperform the vanilla PromptEOL.
Therefore, we contend that the improvements brought by \name\ are not merely due to the reduction in the dimensionality.

(2) \name\ provides the most effective strategy for subspace filtering.
Our comparisons include configuration \circledNum{3} through \circledNum{5}, where we selectively filter the right singular subspaces associated with the largest~(Dominant), smallest~(Secondary), and  intermediate~(Bulk) singular values, respectively.
Compared to these variants, \name\ achieves the best downstream performance.
Notably, configuration \circledNum{5} --- the inverse operation of \name\ --- obtains the poorest results. 
Moreover, we find that Configuration \circledNum{4} significantly outperforms \circledNum{5}.
This finding is in line with the \(\Delta \pi\) distribution shown in Figure~\ref{fig:edge}, where the secondary subspace exhibits a greater tendency to encode frequent tokens than the dominant subspace. 
We leave the exploration of optimal strategies for filtering the asymmetric edge spectrum subspace to future work.

(3) \name\ is remarkably effective, nearly reaching the theoretical upper bound of our framework's potential. In configuration \circledNum{6}, we identify singular vectors with the largest $\Delta \pi^{\mathrm{(i)}}$ based on our analysis in Section~\ref{sec:interpret} and filter out the corresponding subspace. 
We regard this configuration as the theoretical upper bound of \name's capability. As shown in Table~\ref{tab:ablation}, \name\ performs competitively with configuration \circledNum{6} while requiring no task-specific calibration and being significantly simpler to implement.

\subsection{Comparison between \name and  Embedding Calibration Baselines}
We also compare \name\ with established embedding calibration baselines.
These methods typically derive text embeddings from a calibration dataset and propose improvements based on the resulting statistical properties.
A representative baseline is Bert-whitening~\cite{whitening}, which addresses the anisotropic issue by applying a whitening operation to the text embeddings.
Notably, BERT-whitening also facilitates dimensionality reduction consequently.

Given this, we compare \name~and whitening on Qwen and set \(\tau=2\).
We follow the experimental setups from~\cite{whitening}, and report the results with supervision of NLI dataset~\cite{allni}.
Their results on MTEB are presented in Table~\ref{tab:whitening}.
While whitening helps improve the performance, \name~still outperforms it without the supervision of any calibration data.
We argue that the unembedding matrix of LLMs captures valuable statistical features during the pretraining phase that have been previously overlooked.
We did not include this method as a baseline in Table~\ref{tab:main_results}, as its reliance on calibration data would lead to an unfair comparison with \name.

While \name\ is primarily heuristic, we also provide a whitening perspective to help understand. 
In effect, it can be interpreted as a whitening-like operation within bulk spectral space:
{\Large
\begin{equation*}
    \widetilde{\ve}_i \; = \; \ve_i \, \bm{\Phi}_r^\top \; = \; \sum_{j=l_\tau}^{r_\tau} \alpha_j \, \vv_j, \quad \text{where} \;\, \alpha_j \; = \; \operatorname{proj}_{\vv_j} \ve_i.
\end{equation*}
}
Text embeddings exhibit more uniform projections onto directions associated with mid-range singular values, providing a relatively isotropic subspace for free.
We leave a deeper investigation into the underlying mechanisms of this phenomenon to future work, and we hope this perspective will inspire readers and inform future advancements in text embedding training.

\section{Conclusion}
In this paper, we investigate the suboptimal zero-shot performance of LLMs on text embedding tasks and provide a mechanistic interpretation.
Through an analysis of the model's unembedding matrix, we discover the edge spectrum space, which is responsible for encoding high-frequency tokens into the embedding space.
Motivated by this finding, we introduce \name, a simple linear transformation to filter out this spectrum space.
Our experiments across multiple LLM backbones demonstrate that applying \name\ leads to superior zero-shot improvements on text embedding tasks. 
Crucially, we also find that this filtering design implicitly reduces the effective dimensionality of the embeddings, thereby lowering index storage overhead and accelerating retrieval.
We hope our findings provide insights and inspire more principled designs to improve text embeddings training.

\section*{Acknowledgment}
This work is supported by Lenovo Group.
We thank Ang Lv for his writing suggestions and guidance during the rebuttal phase. 
We are also grateful to Yuhan Liu and Yankai Lin for providing computational resources and API access. 
Additionally, we sincerely acknowledge the anonymous KDD reviewers for their constructive comments and questions, which have greatly improved this work.

\bibliographystyle{ACM-Reference-Format}
\balance
\bibliography{sample-base}

\appendix

\clearpage
\newpage

\section{Details of the Main Experimental Setup}
\label{sec:main_setup}
In this section, we provide additional details about the experimental setups discussed in Section~\ref{sec:exp}.
We evaluate all tasks from MTEB, including semantic textual similarity (STS.), classification (Class.), clustering (Cluster.), pair classification (PairClass.), re-ranking (Rerank.), retrieval (Retr.), and summarization (Sum.). 
Due to limited computational resources, we evaluate a subset of the retrieval tasks, consisting of eight datasets~\cite{mteb}: SciFact~\cite{scifact}, ArguAna~\cite{arguana}, NFCorpus~\cite{Nfcorpus}, FiQA2018~\cite{fiqa}, QuoraRetrieval~\cite{beir}, SCIDOCS~\cite{scidocs}, Touche2020~\cite{touche}, TRECCOVID~\cite{treccovid}. 
Finally we use the metrics recommended by MTEB, showing in Table~\ref{tab:mteb_metrics}, where the Spearman’s correlation is calculated on cosine similarity. 
For \name\ used on Mistral-7B-Instruct-V0.3, we offset the whole indices until \(l_\tau=128\).
We provide the actual prompts used for PromptEOL and ECHO across different models below; "text" denotes the sentences to be embedded.
\begin{tcolorbox}[
    colback=white,
    arc=0mm,
    title=PromptEOL,
    boxrule=0.5pt, 
    left = 0mm, right = 0mm,
    top=1mm, bottom=1mm
]
    \label{box:PromptEOL}
    \textbf{Qwen}
    \begin{tcolorbox}[colback=black!3, boxrule=0.2pt, arc=1mm, top=0.2mm, bottom=0.2mm, left = 0mm, right = 0mm, before skip=2mm, after skip=2mm]
        Summarize the sentence: "\{text\}" in one word:"
    \end{tcolorbox}

    \textbf{Llama}
    \begin{tcolorbox}[colback=black!3, boxrule=0.2pt, arc=1mm, top=0.2mm, bottom=0.2mm, left = 0mm, right = 0mm, before skip=2mm, after skip=2mm]
        Summarize the sentence: "\{text\}" in one word:"
    \end{tcolorbox}
    
    \textbf{Mistral}
    \begin{tcolorbox}[colback=black!3, boxrule=0.2pt, arc=1mm, top=0.2mm, bottom=0.2mm, left = 0mm, right = 0mm, before skip=2mm, after skip=2mm]
        This sentence: "\{text\}" means [MASK]
    \end{tcolorbox}
\end{tcolorbox}

\vspace{0.5mm}

\begin{tcolorbox}[
    colback=white,
    arc=0mm,
    title=ECHO,
    boxrule=0.5pt, 
    left = 0mm, right = 0mm,
    top=0.2mm, bottom=0.2mm,
    fontupper=\small
]
    \label{box:ECHO}
    \textbf{Qwen}
    \begin{tcolorbox}[colback=black!3, boxrule=0.2pt, arc=1mm, top=0.2mm, bottom=0.2mm, left = 0mm, right = 0mm, before skip=2mm, after skip=2mm]
        Rewrite the following paragraph: \{text\}. The rewritten paragraph: \{text\}
    \end{tcolorbox}

    \textbf{Llama}
    \begin{tcolorbox}[colback=black!3, boxrule=0.2pt, arc=1mm, top=0.2mm, bottom=0.2mm, left = 0mm, right = 0mm, before skip=2mm, after skip=2mm]
        Rewrite the following paragraph: \{text\}. The rewritten paragraph: \{text\}
    \end{tcolorbox}
    
    \textbf{Mistral}
    \begin{tcolorbox}[colback=black!3, boxrule=0.2pt, arc=1mm, top=0.2mm, bottom=0.2mm, left = 0mm, right = 0mm, before skip=2mm, after skip=2mm]
        Rewrite the following paragraph: \{text\}. The rewritten paragraph: \{text\}
    \end{tcolorbox}
\end{tcolorbox}

\begin{table}[h]
\centering
\small
\vskip -0.05in
\caption{Evaluation metrics used for MTEB tasks.}
\label{tab:mteb_metrics}
\resizebox{0.86\columnwidth}{!}{
\begin{tabular}{ll}
\toprule
\textbf{Task Category} & \textbf{Evaluation Metric} \\
\midrule
STS & Spearman’s correlation \\
Classification & Accuracy \\
Clustering & V-measure \\
Pair Classification & Average Precision (AP) \\
Reranking & Mean Average Precision (MAP) \\
Retrieval & nDCG@10 \\
Summarization & Spearman’s correlation \\
\bottomrule
\end{tabular}
}
\end{table}


    


\section{Equivalence Transformation Proof}
\label{sec:proof}
In the main text, we define the projection matrix as:
{\Large
\[
    \Phi_{\tau} = \mV[l_{\tau} : r_{\tau}] \, \mV[l_{\tau} : r_{\tau}]^{\top}.
\]
}
Let $\mV_\tau = \mV[l_{\tau} : r_{\tau}]$ for simplicity, we seek to prove the identity:
{\Large
\begin{equation*}
    \|\vx \, \bm{\Phi_{\tau}}^{\top} - \vy \, \bm{\Phi_{\tau}}^{\top}\|_2 = \|\vx \, \mV_\tau - \vy \, \mV_\tau\|_2.
\end{equation*}
}
Let $\vz = \vx - \vy$. 
The left-hand side can be written as:
{\Large
\[
    \|\vx \, \bm{\Phi_{\tau}}^{\top} - \vy \, \bm{\Phi_{\tau}}^{\top}\|_2 \; = \; \|\vz \, \bm{\Phi_{\tau}}^{\top}\|_2 \; = \; \|\vz \, \mV_\tau \, \mV_\tau^{\top}\|_2,
\]
}
considering that $\mV_\tau \, \mV_\tau^{\top}$ is identity, thus we have:
{\Large
\[
     \|\vz \, \mV_\tau \, \mV_\tau^{\top}\|_2 \; = \; \|\vz\|_2 \; = \; \|\vx - \vy\|_2.
\]
}
this completes the proof of the identity in \eqref{eq:identity}.

\begin{figure}[b]
    \centering
    \includegraphics[width=\linewidth]{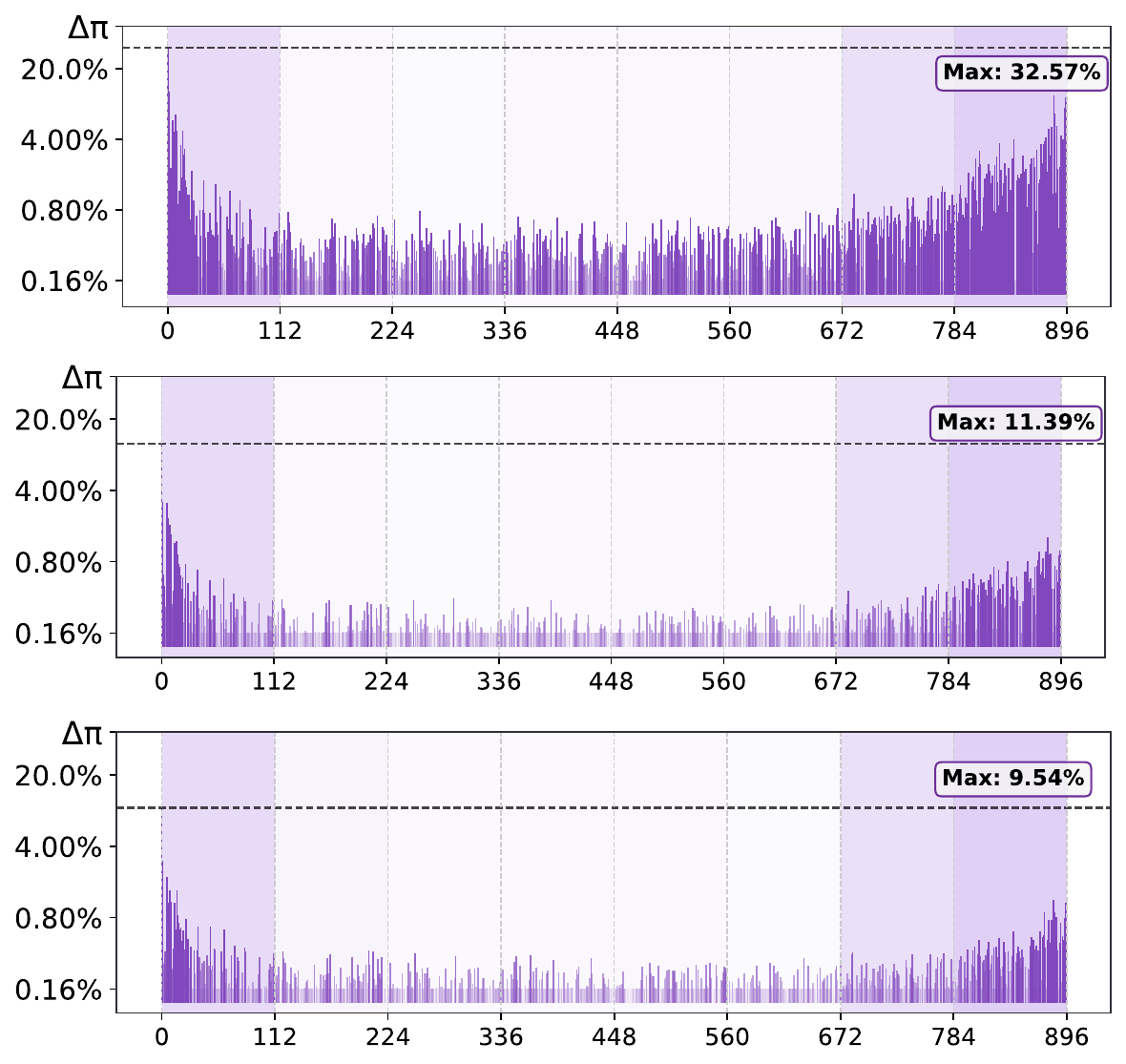}
    \caption{\(\Delta \pi\) distribution for high-frequency, low-frequency and randomly sampled tokens on the Qwen model.}
    \label{fig:infreq}
\end{figure}

\end{document}

%% file: math_commands.tex
\usepackage{amsmath,amsfonts,bm}

\def\eqref#1{equation~\ref{#1}}

\def\1{\bm{1}}

\def\vb{{\bm{b}}}

\def\ve{{\bm{e}}}

\def\vh{{\bm{h}}}

\def\vp{{\bm{p}}}
\def\vq{{\bm{q}}}

\def\vv{{\bm{v}}}
\def\vw{{\bm{w}}}
\def\vx{{\bm{x}}}
\def\vy{{\bm{y}}}
\def\vz{{\bm{z}}}

\def\mI{{\bm{I}}}

\def\mU{{\bm{U}}}
\def\mV{{\bm{V}}}
\def\mW{{\bm{W}}}
\def\mX{{\bm{X}}}

\DeclareMathAlphabet{\mathsfit}{\encodingdefault}{\sfdefault}{m}{sl}
\SetMathAlphabet{\mathsfit}{bold}{\encodingdefault}{\sfdefault}{bx}{n}



%% file: sample-base.bib
@String{Springer = "Springer-Verlag" }

@misc{r1,
      title={DeepSeek-V4: Towards Highly Efficient Million-Token Context Intelligence},
      author={DeepSeek-AI},
      year={2026},
}

@article{scaling,
  title={Scaling laws for neural language models},
  author={Kaplan, Jared and McCandlish, Sam and Henighan, Tom and Brown, Tom B and Chess, Benjamin and Child, Rewon and Gray, Scott and Radford, Alec and Wu, Jeffrey and Amodei, Dario},
  journal={arXiv preprint arXiv:2001.08361},
  year={2020}
}

@inproceedings{prompteol,
    title = "Scaling Sentence Embeddings with Large Language Models",
    author = "Jiang, Ting  and
      Huang, Shaohan  and
      Luan, Zhongzhi  and
      Wang, Deqing  and
      Zhuang, Fuzhen",
    editor = "Al-Onaizan, Yaser  and
      Bansal, Mohit  and
      Chen, Yun-Nung",
    booktitle = "Findings of the Association for Computational Linguistics: EMNLP 2024",
    month = nov,
    year = "2024",
    address = "Miami, Florida, USA",
    publisher = "Association for Computational Linguistics",
    url = "https://aclanthology.org/2024.findings-emnlp.181/",
    doi = "10.18653/v1/2024.findings-emnlp.181",
    pages = "3182--3196"
}

@inproceedings{
    echo,
    title={Repetition Improves Language Model Embeddings},
    author={Jacob Mitchell Springer and Suhas Kotha and Daniel Fried and Graham Neubig and Aditi Raghunathan},
    booktitle={The Thirteenth International Conference on Learning Representations},
    year={2025},
    url={https://openreview.net/forum?id=Ahlrf2HGJR}
}

@inproceedings{
    moee,
    title={Your Mixture-of-Experts {LLM} Is Secretly an Embedding Model for Free},
    author={Ziyue Li and Tianyi Zhou},
    booktitle={The Thirteenth International Conference on Learning Representations},
    year={2025},
    url={https://openreview.net/forum?id=eFGQ97z5Cd}
}

@misc{llm2vec,
      title={LLM2Vec: Large Language Models Are Secretly Powerful Text Encoders}, 
      author={Parishad BehnamGhader and Vaibhav Adlakha and Marius Mosbach and Dzmitry Bahdanau and Nicolas Chapados and Siva Reddy},
      year={2024},
      eprint={2404.05961},
      archivePrefix={arXiv},
      primaryClass={cs.CL},
      url={https://arxiv.org/abs/2404.05961}, 
}

@inproceedings{metaeol,
    title = "Meta-Task Prompting Elicits Embeddings from Large Language Models",
    author = "Lei, Yibin  and
      Wu, Di  and
      Zhou, Tianyi  and
      Shen, Tao  and
      Cao, Yu  and
      Tao, Chongyang  and
      Yates, Andrew",
    editor = "Ku, Lun-Wei  and
      Martins, Andre  and
      Srikumar, Vivek",
    booktitle = "Proceedings of the 62nd Annual Meeting of the Association for Computational Linguistics (Volume 1: Long Papers)",
    month = aug,
    year = "2024",
    address = "Bangkok, Thailand",
    publisher = "Association for Computational Linguistics",
    url = "https://aclanthology.org/2024.acl-long.546/",
    doi = "10.18653/v1/2024.acl-long.546",
    pages = "10141--10157"
}

@inproceedings{geneol,
    title = "{G}en{EOL}: Harnessing the Generative Power of {LLM}s for Training-Free Sentence Embeddings",
    author = "Thirukovalluru, Raghuveer  and
      Dhingra, Bhuwan",
    editor = "Chiruzzo, Luis  and
      Ritter, Alan  and
      Wang, Lu",
    booktitle = "Findings of the Association for Computational Linguistics: NAACL 2025",
    month = apr,
    year = "2025",
    address = "Albuquerque, New Mexico",
    publisher = "Association for Computational Linguistics",
    url = "https://aclanthology.org/2025.findings-naacl.122/",
    doi = "10.18653/v1/2025.findings-naacl.122",
    pages = "2295--2308",
    ISBN = "979-8-89176-195-7"
}

@article{logitlens,
  title={Eliciting Latent Predictions from Transformers with the Tuned Lens},
  author={Belrose, Nora and Furman, Zach and Smith, Logan and Wu, Jason and Ge, Brian and Trakhtenberg, Alisa and Shah, Misha and Gurney, Jacob},
  journal={arXiv preprint arXiv:2303.08112},
  year={2023}
}

@inproceedings{bertflow,
  title={On the Sentence Embeddings from Pre-trained Language Models},
  author={Li, Bohan and Zhou, Hao and He, Junxian and Wang, Mingxuan and Yang, Yiming and Li, Lei},
  booktitle={Proceedings of the 2020 Conference on Empirical Methods in Natural Language Processing (EMNLP)},
  pages={9119--9130},
  year={2020}
}

@inproceedings{ethayarajh-2019-contextual,
    title = "How Contextual are Contextualized Word Representations? {C}omparing the Geometry of {BERT}, {ELM}o, and {GPT}-2 Embeddings",
    author = "Ethayarajh, Kawin",
    editor = "Inui, Kentaro  and
      Jiang, Jing  and
      Ng, Vincent  and
      Wan, Xiaojun",
    booktitle = "Proceedings of the 2019 Conference on Empirical Methods in Natural Language Processing and the 9th International Joint Conference on Natural Language Processing (EMNLP-IJCNLP)",
    month = nov,
    year = "2019",
    address = "Hong Kong, China",
    publisher = "Association for Computational Linguistics",
    url = "https://aclanthology.org/D19-1006/",
    doi = "10.18653/v1/D19-1006",
    pages = "55--65"
}

@misc{lv2024fact,
      title={Interpreting Key Mechanisms of Factual Recall in Transformer-Based Language Models}, 
      author={Ang Lv and Yuhan Chen and Kaiyi Zhang and Yulong Wang and Lifeng Liu and Ji-Rong Wen and Jian Xie and Rui Yan},
      year={2024},
      eprint={2403.19521},
      archivePrefix={arXiv},
      primaryClass={cs.CL},
      url={https://arxiv.org/abs/2403.19521}, 
}

@inproceedings{spectral,
    title = "Spectral Filters, Dark Signals, and Attention Sinks",
    author = "Cancedda, Nicola",
    editor = "Ku, Lun-Wei  and
      Martins, Andre  and
      Srikumar, Vivek",
    booktitle = "Proceedings of the 62nd Annual Meeting of the Association for Computational Linguistics (Volume 1: Long Papers)",
    month = aug,
    year = "2024",
    address = "Bangkok, Thailand",
    publisher = "Association for Computational Linguistics",
    url = "https://aclanthology.org/2024.acl-long.263/",
    doi = "10.18653/v1/2024.acl-long.263",
    pages = "4792--4808"
}

@inproceedings{nie2024text,
    title = "A Text is Worth Several Tokens: Text Embedding from {LLM}s Secretly Aligns Well with The Key Tokens",
    author = "Nie, Zhijie  and
      Zhang, Richong  and
      Wu, Zhanyu",
    editor = "Che, Wanxiang  and
      Nabende, Joyce  and
      Shutova, Ekaterina  and
      Pilehvar, Mohammad Taher",
    booktitle = "Proceedings of the 63rd Annual Meeting of the Association for Computational Linguistics (Volume 1: Long Papers)",
    month = jul,
    year = "2025",
    address = "Vienna, Austria",
    publisher = "Association for Computational Linguistics",
    url = "https://aclanthology.org/2025.acl-long.379/",
    doi = "10.18653/v1/2025.acl-long.379",
    pages = "7683--7694",
    ISBN = "979-8-89176-251-0"
}

@misc{whitening,
      title={Whitening Sentence Representations for Better Semantics and Faster Retrieval}, 
      author={Jianlin Su and Jiarun Cao and Weijie Liu and Yangyiwen Ou},
      year={2021},
      eprint={2103.15316},
      archivePrefix={arXiv},
      primaryClass={cs.CL},
      url={https://arxiv.org/abs/2103.15316}, 
}

@misc{qwen2.5,
    title = {Qwen2.5: A Party of Foundation Models},
    url = {https://qwenlm.github.io/blog/qwen2.5/},
    author = {Qwen Team},
    month = {September},
    year = {2024}
}

@article{llama3,
  title={The llama 3 herd of models},
  author={Grattafiori, Aaron and Dubey, Abhimanyu and Jauhri, Abhinav and Pandey, Abhinav and Kadian, Abhishek and Al-Dahle, Ahmad and Letman, Aiesha and Mathur, Akhil and Schelten, Alan and Vaughan, Alex and others},
  journal={arXiv preprint arXiv:2407.21783},
  year={2024}
}

@misc{mistral7b,
      title={Mistral 7B}, 
      author={Albert Q. Jiang and Alexandre Sablayrolles and Arthur Mensch and Chris Bamford and Devendra Singh Chaplot and Diego de las Casas and Florian Bressand and Gianna Lengyel and Guillaume Lample and Lucile Saulnier and Lélio Renard Lavaud and Marie-Anne Lachaux and Pierre Stock and Teven Le Scao and Thibaut Lavril and Thomas Wang and Timothée Lacroix and William El Sayed},
      year={2023},
      eprint={2310.06825},
      archivePrefix={arXiv},
      primaryClass={cs.CL},
      url={https://arxiv.org/abs/2310.06825}, 
}

@misc{redpajama,
      title={RedPajama: an Open Dataset for Training Large Language Models}, 
      author={Maurice Weber and Daniel Fu and Quentin Anthony and Yonatan Oren and Shane Adams and Anton Alexandrov and Xiaozhong Lyu and Huu Nguyen and Xiaozhe Yao and Virginia Adams and Ben Athiwaratkun and Rahul Chalamala and Kezhen Chen and Max Ryabinin and Tri Dao and Percy Liang and Christopher Ré and Irina Rish and Ce Zhang},
      year={2024},
      eprint={2411.12372},
      archivePrefix={arXiv},
      primaryClass={cs.CL},
      url={https://arxiv.org/abs/2411.12372}, 
}

@article{inverse,
  author  = {Penrose, R.},
  title   = {A generalized inverse for matrices},
  journal = {Proceedings of the Cambridge Philosophical Society},
  year    = {1955},
  volume  = {51},
  number  = {3},
  pages   = {406--413},
  doi     = {10.1017/S0305004100030784}
}

@inproceedings{mteb,
    title = "{MTEB}: Massive Text Embedding Benchmark",
    author = "Muennighoff, Niklas  and
      Tazi, Nouamane  and
      Magne, Loic  and
      Reimers, Nils",
    editor = "Vlachos, Andreas  and
      Augenstein, Isabelle",
    booktitle = "Proceedings of the 17th Conference of the European Chapter of the Association for Computational Linguistics",
    month = may,
    year = "2023",
    address = "Dubrovnik, Croatia",
    publisher = "Association for Computational Linguistics",
    url = "https://aclanthology.org/2023.eacl-main.148/",
    doi = "10.18653/v1/2023.eacl-main.148",
    pages = "2014--2037"
}

@article{kusupati2022matryoshka,
  title={Matryoshka representation learning},
  author={Kusupati, Aditya and Bhatt, Gantavya and Rege, Aniket and Wallingford, Matthew and Sinha, Aditya and Ramanujan, Vivek and Howard-Snyder, William and Chen, Kaifeng and Kakade, Sham and Jain, Prateek and others},
  journal={Advances in Neural Information Processing Systems},
  volume={35},
  pages={30233--30249},
  year={2022}
}

@inproceedings{simcse,
    title = "{S}im{CSE}: Simple Contrastive Learning of Sentence Embeddings",
    author = "Gao, Tianyu  and
      Yao, Xingcheng  and
      Chen, Danqi",
    editor = "Moens, Marie-Francine  and
      Huang, Xuanjing  and
      Specia, Lucia  and
      Yih, Scott Wen-tau",
    booktitle = "Proceedings of the 2021 Conference on Empirical Methods in Natural Language Processing",
    month = nov,
    year = "2021",
    address = "Online and Punta Cana, Dominican Republic",
    publisher = "Association for Computational Linguistics",
    url = "https://aclanthology.org/2021.emnlp-main.552/",
    doi = "10.18653/v1/2021.emnlp-main.552",
    pages = "6894--6910"
}

@inproceedings{cocondensor,
    title = "Unsupervised Corpus Aware Language Model Pre-training for Dense Passage Retrieval",
    author = "Gao, Luyu  and
      Callan, Jamie",
    editor = "Muresan, Smaranda  and
      Nakov, Preslav  and
      Villavicencio, Aline",
    booktitle = "Proceedings of the 60th Annual Meeting of the Association for Computational Linguistics (Volume 1: Long Papers)",
    month = may,
    year = "2022",
    address = "Dublin, Ireland",
    publisher = "Association for Computational Linguistics",
    url = "https://aclanthology.org/2022.acl-long.203/",
    doi = "10.18653/v1/2022.acl-long.203",
    pages = "2843--2853"
}

@inproceedings{allni,
    title = "A large annotated corpus for learning natural language inference",
    author = "Bowman, Samuel R.  and
      Angeli, Gabor  and
      Potts, Christopher  and
      Manning, Christopher D.",
    editor = "M{\`a}rquez, Llu{\'\i}s  and
      Callison-Burch, Chris  and
      Su, Jian",
    booktitle = "Proceedings of the 2015 Conference on Empirical Methods in Natural Language Processing",
    month = sep,
    year = "2015",
    address = "Lisbon, Portugal",
    publisher = "Association for Computational Linguistics",
    url = "https://aclanthology.org/D15-1075",
    doi = "10.18653/v1/D15-1075",
    pages = "632--642",
}

@inproceedings{arguana,
  author = {Wachsmuth, Henning and Syed, Shahbaz and Stein, Benno},
  booktitle = {ACL},
  title = {Retrieval of the Best Counterargument without Prior Topic Knowledge},
  year = {2018},
}

@inproceedings{scifact,
    title = "Fact or Fiction: Verifying Scientific Claims",
    author = "Wadden, David  and
      Lin, Shanchuan  and
      Lo, Kyle  and
      Wang, Lucy Lu  and
      van Zuylen, Madeleine  and
      Cohan, Arman  and
      Hajishirzi, Hannaneh",
    booktitle = "Proceedings of the 2020 Conference on Empirical Methods in Natural Language Processing (EMNLP)",
    month = nov,
    year = "2020",
    address = "Online",
    publisher = "Association for Computational Linguistics",
    url = "https://aclanthology.org/2020.emnlp-main.609",
    doi = "10.18653/v1/2020.emnlp-main.609",
    pages = "7534--7550",
}

@inproceedings{Nfcorpus,
  title="A Full-Text Learning to Rank Dataset for Medical Information Retrieval",
  author = "Vera Boteva and Demian Gholipour and Artem Sokolov and Stefan Riezler",
  booktitle = "Proceedings of the European Conference on Information Retrieval ({ECIR})",
  location = "Padova, Italy",
  publisher = "Springer",
  year = 2016
}

@article{fiqa,
  title={WWW'18 Open Challenge: Financial Opinion Mining and Question Answering},
  author={Maia, Macedo and Handschuh, Siegfried and Freitas, Andr{\'e} and Davis, Brian and McDermott, Ross and Zarrouk, Manel and Balahur, Alexandra},
  booktitle={Companion Proceedings of the The Web Conference 2018},
  pages={1941--1942},
  year={2018}
}

@inproceedings{beir,
    title={{BEIR}: A Heterogeneous Benchmark for Zero-shot Evaluation of Information Retrieval Models},
    author={Nandan Thakur and Nils Reimers and Andreas R{\"u}ckl{\'e} and Abhishek Srivastava and Iryna Gurevych},
    booktitle={Thirty-fifth Conference on Neural Information Processing Systems Datasets and Benchmarks Track (Round 2)},
    year={2021},
    url={https://openreview.net/forum?id=wCu6T5xFjeJ}
}

@inproceedings{scidocs,
  author = {Arman Cohan and Sergey Feldman and Iz Beltagy and Doug Downey and Daniel S. Weld},
  booktitle = {ACL},
  title = {SPECTER: Document-level Representation Learning using Citation-informed Transformers},
  year = {2020},
}

@inproceedings{touche,
    author = {Bondarenko, Alexander and Fr\"{o}be, Maik and Beloucif, Meriem and Gienapp, Lukas and Ajjour, Yamen and Panchenko, Alexander and Biemann, Chris and Stein, Benno and Wachsmuth, Henning and Potthast, Martin and Hagen, Matthias},
    title = {Overview of Touch\'{e} 2020: Argument Retrieval: Extended Abstract},
    year = {2020},
    isbn = {978-3-030-58218-0},
    publisher = {Springer-Verlag},
    address = {Berlin, Heidelberg},
    url = {https://doi.org/10.1007/978-3-030-58219-7_26},
    doi = {10.1007/978-3-030-58219-7_26},
    booktitle = {Experimental IR Meets Multilinguality, Multimodality, and Interaction: 11th International Conference of the CLEF Association, CLEF 2020, Thessaloniki, Greece, September 22–25, 2020, Proceedings},
    pages = {384–395},
    numpages = {12},
    location = {Thessaloniki, Greece}
}

@misc{treccovid,
  archiveprefix = {arXiv},
  author = {Kirk Roberts and Tasmeer Alam and Steven Bedrick and Dina Demner-Fushman and Kyle Lo and Ian Soboroff and Ellen Voorhees and Lucy Lu Wang and William R Hersh},
  eprint = {2104.09632},
  primaryclass = {cs.IR},
  title = {Searching for Scientific Evidence in a Pandemic: An Overview of TREC-COVID},
  year = {2021},
}
